# Active Learning for Structured Prediction from Partially Labeled Data

Mehran Khodabandeh, Zhiwei Deng, Mostafa S. Ibrahim, Shin'ichi Satoh, Greg Mori
Simon Fraser University
Burnaby, BC, Canada
{mkhodaba, zhiweid, msibrahi}@sfu.ca, satoh@nii.ac.jp mori@cs.sfu.ca

## Abstract

*We propose a general purpose active learning algorithm for structured prediction – gathering labeled data for training a model that outputs a set of related labels for an image/video. Active learning starts with a limited initial training set, then iterates querying a user for labels on unlabeled data and retraining the model. We propose a novel algorithm for selecting data for labeling, choosing examples to maximize expected information gain based on belief propagation inference. This is a general purpose method and can be applied to a variety of tasks/models. As a specific example we demonstrate this framework for learning to recognize human actions and group activities in video sequences. Experiments show that our proposed algorithm outperforms previous active learning methods and can achieve accuracy comparable to fully supervised methods while utilizing significantly less labeled data.*

## 1. Introduction

Consider the activity recognition problem depicted in Fig. 1. Gathering large quantities of labeled data can lead to accurate classifiers that can predict what each person is doing and the overarching activity taking place in a scene.

However, naive approaches for labeling training data would be inefficient. Video data have significant redundancy; it seems unnecessary to label every single person in each frame of each video. Further, action categories vary significantly in frequency and intra-class variation. If images were selected randomly for labeling, copious quantities of similar walking poses would likely result. Finally, contextual inference that utilizes relationships between people in a scene should be accounted for when deciding whether a sample is necessary for learning.

For these reasons, we focus on active learning for building labeled datasets in structured prediction. In our framework the active learner benefits from partially labeled data. For example, in the task of activity recognition, we have video frames consisting of multiple people with both individual action and overarching activity labels. Our proposed algorithm selects which people and scenes would be most informative to label. This allows labeling effort to hone in on the most important examples – such as unusual poses, categories with intra-class variation, key people in scenes, or ambiguous overarching scene labels.

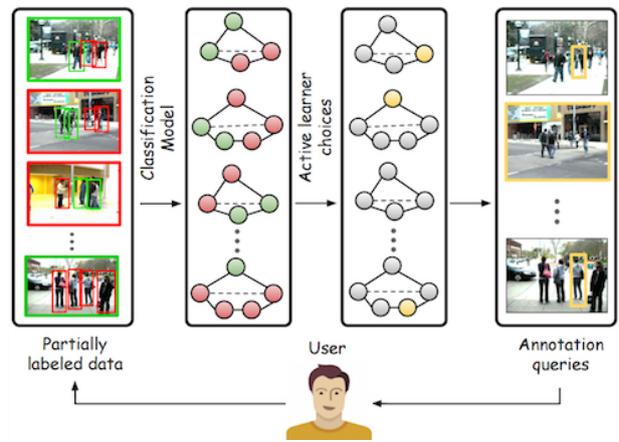

Figure 1: Example of active learning for recognizing human actions. Given a partially labeled dataset (labeled examples in green, unlabeled red), we train a classifier to recognize the actions of each person and group in a scene. Our active learning criterion selects individual people or scenes to be annotated by the user (shown in yellow). These new data are added to the training dataset, and the process repeated.

While we demonstrate our active learning algorithm on human activity recognition, we note that structured prediction is a common task in computer vision. A variety of methods and algorithms have been developed in this vein. Recently, a number of approaches encode graphical model-style structured prediction inference within modern deep neural networks [40, 35, 11, 5]. Our algorithm follows in this line, showing that a novel criterion for active learning can result in highly effective inference for structured prediction with these state of the art techniques.

Active learning has deep roots within the vision/learning literature. The key ingredient of an active learning algo-rithm is the selection strategy used to choose which unlabeled examples to label. Although active learning has substantial literature, there has been limited work on active learning for structured prediction. Standard methods include uncertainty based sampling [20, 19], margin based sampling [28], expected model change [30, 37], and query-by-committee [31]. For example, a typical uncertainty based approach [20] would calculate the entropy of all the people in a scene separately and choose the ones with the highest entropy.

We argue that considering the relationships between nodes in a graph for predicting the probability distribution of the nodes can lead to significant improvements in the selection strategy. In this paper, we propose a novel active learning criterion that exploits the structure of the graphical model and selects the most informative data based on how much information they provide to the whole graph. More specifically, it estimates how much the average entropy of the graph would be reduced if we "knew" the label of a particular node in the graph. This idea is operationalized within our proposed active learning algorithm.

The main contributions of this paper are (1) introducing a novel expected structured entropy reduction as a sample selection criterion for active learning, (2) developing an inference machine based on deep neural networks for efficiently computing this criterion, and (3) demonstrating the effectiveness of this algorithm for labeling data for human activity recognition.

## 2. Related Work

In this paper we develop a novel approach for active learning in structured prediction, applied to human activity recognition. Each of these areas has seen substantial amounts of previous work. Below, we briefly review closely related previous work in each of these areas.

**Active Learning**: A large number of methods and applications of active learning have been explored. Active learning has been used for many tasks such as image classification [21], categorization of images and objects [14, 12], video segmentation [7, 8], and discovery of human interactions [16]. Another prominent use case is labeling large video or image datasets [39, 4, 38, 25]

There are numerous methods to measure the informativeness of samples. *Entropy* [32] is a standard measure, which has been used in many applications. For example, Holub *et al*. [9] developed a measurement based on minimum expected entropy. Varadarajan *et al*. [36] measure entropy reduction over the whole dataset. An alternate uncertainty criteria is least confidence, where the learner queries the instance that has the lowest probability for its most likely label [29]. Roth and Small [24] use the margin between the two most probable classes as an uncertainty measure. Sun *et al*. [34] estimate the joint distribution in a histogram-based method, however full supervision is needed, which could be expensive.

All these methods require the user/oracle to fully label an instance. However, we are interested in structured prediction tasks. Recent active learning methods that use partially labeled data include Luo *et al*. [20], in which the entropies of the marginal distributions are computed using belief propagation. Vezhnevets *et al*. [37] measured the (discrete) count of changes in labels of the nodes in a CRF, retraining the model for each instance and every possible label, on each iteration. We build on these approaches, by taking into account relations between examples, and select instances based on how much information they can provide to other examples as well. Further, rather than counts, we instead we measure expected entropy change which is a more informative, continuous value, within a more efficient belief propagation paradigm.

**Structured Prediction**: Many computer vision problems involve a set of output variables with relations among them. Structured prediction has a long history and recently has been addressed using deep learning. Zheng *et al*. [40] design a network called CRF-RNN which is basically a CRF that is formulated with Guassian pairwise potentials, trainable in and end-to-end manner. Tompson *et al*. [35] build a model consisting of a MRF and a CNN that are jointly trained to exploit relationships between body parts for human pose estimation. Deng *et al*. [6] propose to represent factor graph in deep learning to model group activities. Jain *et al*. [11] use a generic method to train spatio-temporal graphs in a deep learning framework. Deng *et al*. [5] propose a sequential inference model in an RNN framework that is motivated by the traditional message passing inference algorithm. They utilize gates on edges between graph nodes to learn the structure of a graph. Our proposed method utilizes Deng *et al*.'s inference RNN-based inference technique.

**Activity Recognition**: In our work we focus on recognizing human actions and overarching group activities taking place in a scene. Previous work on this topic has shown that modeling this problem as a form of graphical model helps in providing contextual information for recognizing indvidual and group activities [1, 18, 26, 2, 27, 23]. Lan *et al*. [18] used latent variables in a max margin framework to find the most discriminative structure for the scene. Amer *et al*. [1] used grouping nodes to achieve the same goal. Shu *et al*. [33] utilized AND-OR graphs to recognize events and assign roles to the engaged people in noisy tracklets. Khamis *et al*. [15] combined track-level and frame-level information for the task of action recognition. Ramanathan et al. [22] use attention models to focus on the key player in sports videos. Deng *et al*. [5] used Recurrent Neural Networks to learn the structure of the people in a scene. In this work we propose a general active learning method for selecting the most informative nodes in structured data. Activity recognition methods can benefit from our approach since annotating videos/images with a full set of action and activity labels is time-consuming.

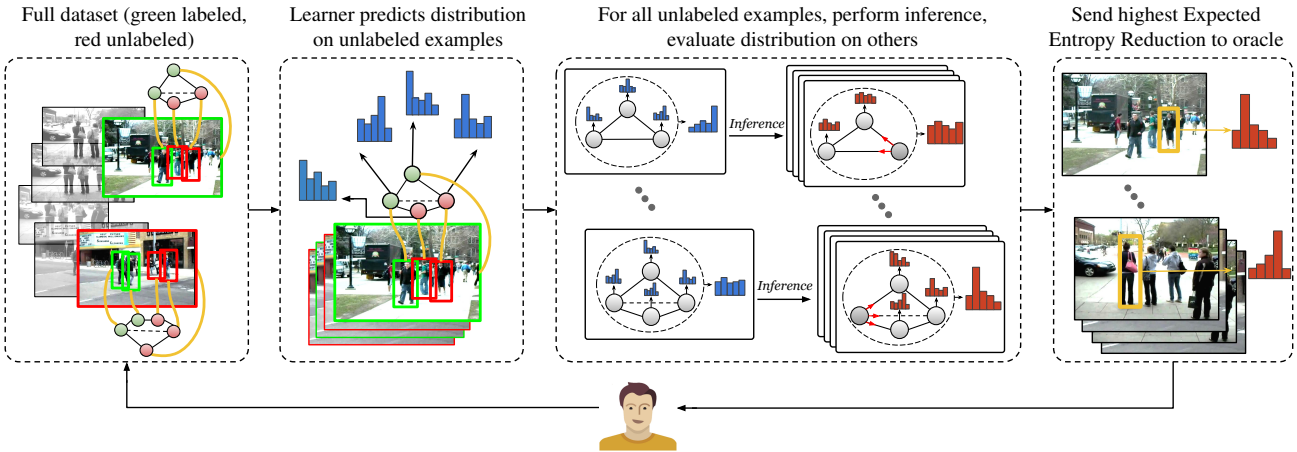

Figure 2: Overview of the method. A structured prediction model is trained given the training dataset. This model produces a probability distribution for every node. In the second stage for each unlabeled node we estimate expected entropy reduction by fixing its value and running inference. We sort all nodes based on expected entropy reduction their labeling would produce. The top $K$ nodes are selected and sent to user for labeling. These nodes can be either human action or scene labels. The new labeled data is then added to the training set and this process repeats.

## 3. Active Learning for Structured Prediction

Successes in visual recognition hinge on the availability of quantities of labeled training data. Acquiring these labeled data efficiently is especially important in human activity recognition. Videos contain much redundant data, and significant structure exists amongst the people present in a scene. Leveraging information regarding the relationships between people and focusing labeling effort on more challenging categories present opportunities for efficiency.

Consider the example images in Fig. 2. Knowing that the action of one person in this scene is *walking* will assist in labeling other people in the scene. As another example, consider the volleyball scenes depicted in Fig. 6. Focusing labeling effort on rare, challenging classes such as *spiking* that help determine the group activity is advantageous.

We formalize our ideas within a typical active learning setting. We start with an initial labeled data set $\mathcal{L}$, and train a model. Next, we select for labeling additional data from an unlabeled pool $\mathcal{U}$. The learner will alternate between these two stages to build a progressively more accurate model. The key question we must answer is: *which unlabeled instances from $\mathcal{U}$ should we choose to label?*

In our work the data are a set of video frames, in which we wish to predict the actions of each person and the overarching group activity taking place. In this structured prediction task, good samples to label are ones about which we are uncertain, and will help to disambiguate many other labels. Hence, our goal is to design an active learning algorithm that annotates the right human actions / group activity labels to improve the learned classifier.

To quantify this we use expected reduction in entropy as our selection criterion. Entropy refers to uncertainty. For example, if an instance has similar probabilities for different labels it has a high entropy. An instance with low entropy means the model can determine the label of that instance with high confidence. We select the most informative instance(s) based on how much entropy is reduced.

This is a simple idea and similar ideas have been used before. However, structured prediction presents new challenges: *how can we know how much labeling a particular unlabeled instance $y_u \in \mathcal{U}$, reduces the entropy?* This is the challenge that we address in this work. In a naive solution, for all possible labels of $y_u$ a classifier is trained on $y_u \cup \mathcal{L}$ and the expected entropy is computed by taking a weighted average. Obviously, this is not feasible because it is time consuming (especially for deep learning) and secondly adding one datum is very noisy.

We address this challenge with an intuitive method based on belief propagation. In order to estimate the entropy if we knew the label of an unlabeled sample $y_u$, we consider it as an observation using every possible value for it, and perform belief propagation. Based on this intuition, for every unlabeled node $y_u$ we calculate the average entropy reduction and select the ones that have the largest entropy reduction. We explain the details of this method below.

### 3.1. Stage 1: Structured Prediction Model

In this section we introduce the formulation of the structured prediction model and how we connect it to the active learning procedure.

We formulate our model as follows. Denote the image observations as $X = \{x_i\}$ and the set of person action / group activity variables we want to model as $Y = \{y_i\}$. We define a graphical model describing the distribution over variables $Y$ conditioned on the observation $X$:

$$P(Y|X;\theta) = \frac{1}{Z_X} \prod_{f \in F} \phi_f(y_f, x_f; \theta_f) \quad (1)$$

in which $\theta$ are the parameters of the model, $F$ is the set of cliques, $y_f$, $x_f$, and $\theta_f$ are variables and parameters of a clique $f$. We utilize graphs that connect each action / group activity label to its corresponding person / scene image features, and a fully connected graph of relations between all people and the group activity in each scene.

In our work we model all distributions using convolutional neural networks. These include potentials that relate image observations to action/group activity labels, and potentials over sets of action/group activity labels in an image.

Inference in this graphical model is conducted using a forward pass through the neural network, which imitates a fixed set of rounds of belief propagation over the graph [5]. In this manner, we can estimate marginalized probability distributions $P(y_i|X;\theta)$ over a particular variable, or likewise $P(y_i|X, y_j = k; \theta)$ given a value for another label $y_j$.

**Parameter Learning.** In active learning, we normally only have a part of the data labeled. Denote the labeled set as $\mathcal{L} = \{y_l\}$. To learn the model parameters $\theta$, we use the loss:

$$\ell(\theta) = -\sum_{y_l \in \mathcal{L}} \log P(y_l = y_l^*|X;\theta) \quad (2)$$

where $y_l^*$ is the provided label for variable $y_l$. This function minimizes the negative log likelihood of the correct classes for the labeled data.

Based on the above formulation, in the next section, we provide the details of how our active learning algorithm chooses new data to be annotated by the oracle.

### 3.2. Stage 2: Instance Selection Strategy

Our goal is to choose informative samples for labeling – samples which will help to reduce ambiguity in labels across entire scenes. The previous section described how we can conduct inference given values for certain nodes. We now utilize this to compute expected entropy reduction to determine the informativeness of samples.

Assume a frame requiring $N$ labels (person actions + group activity) is given; for ease of notation we assume each can take $T$ possible values (actions / group activity). We can obtain the probability distributions $P(y_i|X;\theta)$ for each person and the group activity using the model from Sec. 3.1.

---

**Algorithm 1** Compute Top $K$ Most Informative Instances

1: **procedure** GETINSTANCES(COUNT: INT)
2:    **for all** frame $f$ in dataset **do**
3:       $g \leftarrow$ BUILDGRAPH($f$)
4:       $\Phi^g \leftarrow$ GETEXPECTEDENTROPYREDUCTION($g$)
5:    **end for**
6:    **return** ARGMAXTOPK($\Phi$, $count$)
7:    // *Note:*$\Phi$ is a set of arrays. ArgmaxTopK searches this 2D structure and returns indexes of $K$ largest instances (persons or scenes annotations)
8: **end procedure**

---

**Algorithm 2** Compute Graph Expected Entropy Reduction

1: **procedure** GETEXPECTEDENTROPYREDUCTION(G: GRAPH)
2:    $P \leftarrow$ DOINFERENCE($g$)
3:    $\bar{H} \leftarrow$ GETAVERAGEENTROPY($g, P$)
4:    **for all** node $i$ in graph g **do**
5:       **for** label $j$ in $1..T$ **do**
6:          Fix the label of node $i$ to $j$ by changing probablities
7:          $\hat{P} \leftarrow$ DOINFERENCE($g, i, j$)
8:          $\bar{H}_j^i \leftarrow$ GETAVERAGEENTROPY($g, \hat{P}, i, j$)
9:       **end for**
10:      $E(\bar{H}^i) \leftarrow \sum_{j=1}^{T} P(y_i = j|X;\theta) \bar{H}_j^i$
11:      $\Phi_i^g \leftarrow \bar{H} - E(\bar{H}^i)$
12:    **end for**
13:    **return** $\Phi^g$
14:    // *Note:*$\Phi^g$ is a tuple of nodes computations
15: **end procedure**

---

The entropy, $H_i$, for the label $i$ is defined as:

$$H_i = -\sum_{j=1}^{T} P(y_i = j|X;\theta) \log P(y_i = j|X;\theta) \quad (3)$$

where $P(y_i = j|X;\theta)$ indicates the probability of assigning label $j$ to the variable $i$. The *average entropy*, $\bar{H}$, of a graph with $N$ nodes is the mean of the individual node entropies:

$$\bar{H} = \frac{1}{N} \sum_{i=1}^{N} H_i \quad (4)$$

**Expected average entropy**. In a graph $g$ with some unlabeled nodes, in order to find the most informative node, we need to find the node that reduces the *average entropy* of the graph if its label is known. Computing the actual *average entropy* of the graph if we label an example is not possible. This is because the labels that would be given to examples are not known, and if a certain label is given to an example it impacts the entropy over other nodes in the graph. Therefore, we approximate it with the *expected total entropy*.

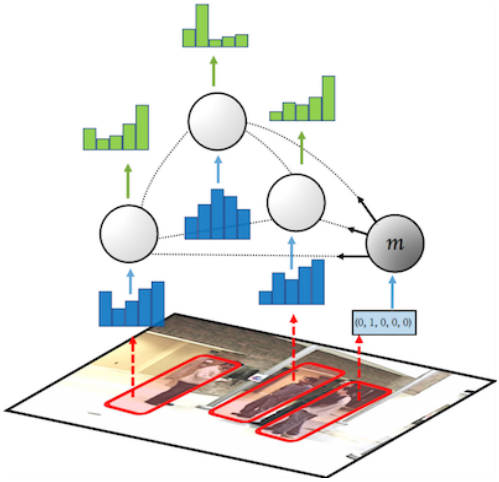

Figure 3: In the above graph node $m$ is unlabeled. In order to compute the *average entropy* of the graph if the label of node $m$ was given and equal to 2 we do the following. First set the label distribution of the node $m$ to $[0, 1, 0, 0, 0]$ ("2"). Then we perform inference using the current learned inference machine. This will produce the probability distributions of all nodes, $\hat{P}$. Then the *average entropy* is the mean entropy over all the nodes. In this figure the blue histograms are the action label distributions of the nodes obtained given the current model; and the green histograms are the probability distributions of the nodes after observing $m = 2$.

The key idea behind the expected average entropy is the following. Suppose we choose to ask a user to label node $i$ in graph $g$. We currently have a belief (probability distribution) over the possible labels the user could give us for this node. For each of these possible resultant labels, we can estimate the average entropy that would remain. This is done by running our structured prediction model while fixing the value of node $i$. The expected average entropy computes a weighted average of these total entropies according to our current belief about node $i$.

In detail, denote by $\bar{H}^i_j$ the average entropy of the graph $g$ if the label of node $i$ is known and is equal to $j$ (symbol $g$ is omitted for simplicity). In order to compute $\bar{H}^i_j$, we fix the label of node $i$ to $j$ and run inference to obtain the probability distributions of all the other nodes. Then we compute the average entropy of the graph using Eqs. 3 and 4.

$$\bar{H}^i_j = -\frac{1}{N} \sum_{\substack{n=1 \\ n \neq i}}^{N} \sum_{t=1}^{T} \hat{P}(y_n = t) \log \hat{P}(y_n = t) \quad (5)$$

where $\hat{P}(y_m = t) \equiv P(y_m = t | X, y_i = j; \theta)$ is the probability of node $m$ having the label $t$, after fixing the label of node $i$ to $j$. We can compute $\bar{H}^i_j$ for all possible values of $j$ (label of the node $i$).

The expected average entropy $\bar{H}^i$ is defined as:

$$E(\bar{H}^i) = \sum_{j=1}^{T} P(y_i = j | X; \theta) \bar{H}^i_j \quad (6)$$

Fig. 3 illustrates this process. As a result, we can determine for each node $i$ in the graph, what we expect to happen if this node were chosen for labeling by the user.

**Expected Average Entropy Reduction.** Finally, we choose to label the node(s) $i$ that would result in the largest reduction in entropy. Denote by $\Phi$ the amount of information that nodes could provide to their graphs if labels are known. For a particular node $i$ in a graph $g$ we define the expected average entropy reduction as:

$$\Phi^g_i = \bar{H} - E(\bar{H}^i) \quad (7)$$

Then, to add the next $K$ training instances $\mathcal{S}$ to our dataset, we get the top $K$ most informative nodes in all available graphs from $\Phi$ (which is a set of arrays, for every graph, $\Phi^g_i$ is computed for its nodes). This can be formulated as the following:

$$\mathcal{S} = \text{ArgMaxTopK}(\Phi, K) \quad (8)$$

where $ArgmaxTopK$ returns the positions of the top $K$ entries (corresponding to person or scene nodes) in $\Phi$.

**Summary.** The complete algorithm is as follows. Given a structured prediction model with parameters $\theta_t$, labeled set $\mathcal{L}_t$, and unlabeled set $\mathcal{U}_t$ at iteration $t$, our active learning process evaluates each node in $\mathcal{U}_t$ by determining how much entropy reduction we expect to obtain by labeling it. We select a set of nodes $\mathcal{C}_t \subseteq \mathcal{U}_t$ for labeling by the user according to this criterion. These nodes are annotated and labeled/unlabeled sets are updated: $\mathcal{L}_{t+1} = \mathcal{L}_t \cup \mathcal{C}_t$; $\mathcal{U}_{t+1} = \mathcal{U}_t \setminus \mathcal{C}_t$. Then the structured prediction model is re-trained and new parameters $\theta_{t+1}$ obtained, and the process repeated. Algorithms 1 and 2 summarize this process. We provide a reference implementation in Caffe to enable reproduction of the results.

## 4. Experiments

We evaluated our method on the task of group activity recognition, which involves structured prediction of individual human actions and group activities. We utilize the Volleyball Dataset [10] and Collective Activity Dataset [3] to evaluate the performance of our method.

Below, we compare to other active learning methods, structured prediction baselines, and approaches based on entropy. We further analyze and interpret the performance of our method. The supplementary material contains details on training settings and results visualizations.

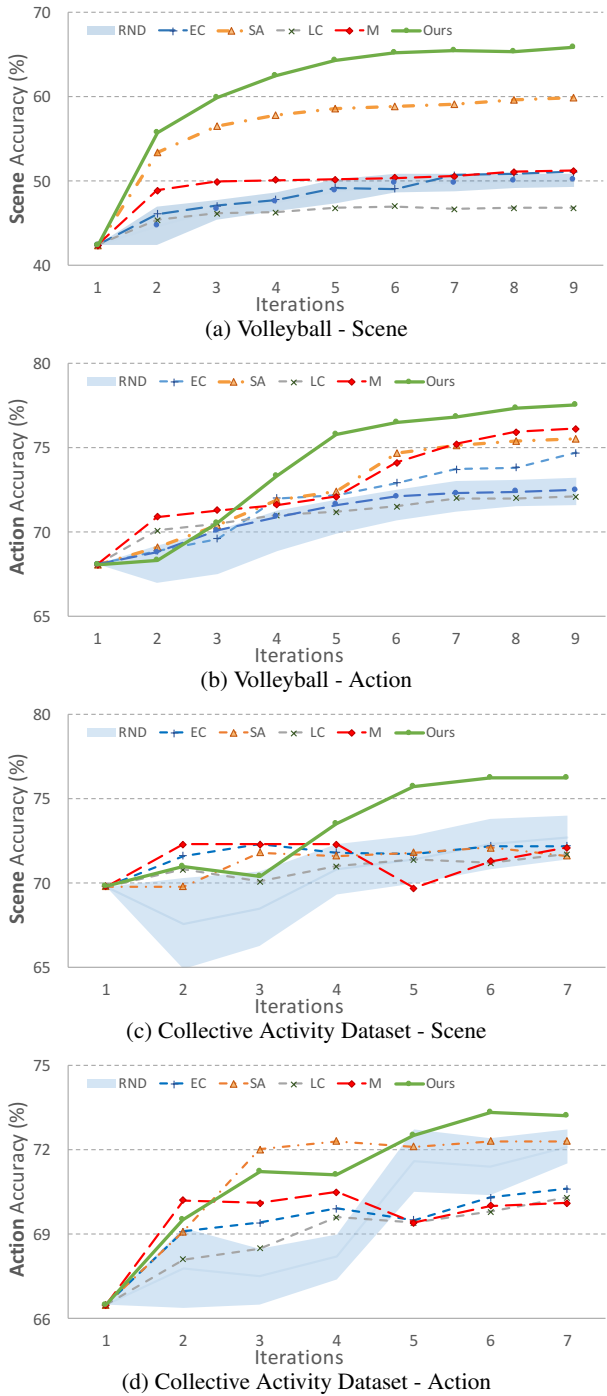

Figure 4: Results of comparison of our method against baselines on **Volleyball Dataset** (a and b), **Collective Activity Dataset** (c and d). The y-axis is accuracy and x-axis is iteration. For all the methods we start from the same small initial labeled set so the accuracies of the first column are exactly the same. Our method outperforms the other methods in both action and scene accuracy[1].

## 4.1. Datasets

**Volleyball Dataset** [10]: This dataset contains 4830 annotated frames from 55 volleyball videos with *nine* player action labels (waiting, setting, digging, falling, spiking, blocking, jumping, moving, standing) and *eight* team activity labels (right set, right spike, right pass, right winpoint, left winpoint, left pass, left spike, left set). The standing action represents around 70% of the action labels, making the dataset very imbalanced. The distribution of *action labels* and *scene labels* are illustrated in the right-most histogram of Figures 5b and 5c respectively. We follow the same datasets split as [10], in which 3493 frames are used for training and 1337 frames for testing. There is a maximum of 12 people in each frame, and the number of people varies from frame to frame. The total number of annotations in the training set is *44,294* (3493 scene labels and 40801 action labels). From the training set, we only use 696 frames for the initial fully labeled training data and the rest ($\approx 35,400$ annotations) is the unlabeled pool.

**Collective Activity Dataset** [3]: This dataset contains 44 videos of outdoor/indoor activities Crossing, Waiting, Talking, Queueing, and Walking. Each individual person could have one of the following 6 labels: Crossing, Waiting, Talking, Queueing, Walking, and Not Available. We use the same train/test split as [18]. The training set contains 1908 frames. The total number of annotations (scene/action labels) in the training set is *11,734*. We only use 382 frames for the initial fully labeled set and the rest ($\approx 9,400$ annotations) is used as the "unlabeled" set.

A model is trained using the initial training set. Our method iteratively selects a number of scenes and queries the user to label certain people in some scenes (a total of $K$ annotations per iteration). The number of people that are labeled at each iteration depends on the available resources that one has. We show results of $K = 1000$ (and $K = 500$ in the supplementary material).

## 4.2. Baselines

In order to verify the effectiveness of our method, we compared our model against baselines that: (1) utilize entropy in a non-structured prediction setting, (2) utilize alternative structured prediction criteria, and (3) standard active learning methods.

- *Separate Active (SA)* [20]: Select $K$ nodes, including person nodes and scene nodes, with the highest entropy. We implement their "Separate active" method using approximate belief propagation for inference.
- *Least Confidence (LC)* [29]: Among all the person/scene nodes in all the graphs (frames), select $K$

---
[1]The accuracy versus added annotation is not a monotonically increasing function. Due to existence of multiple local optima, the accuracy of the model could decrease. Also there is a possibility that the model overfits to the given training set.

nodes that the trained model has the lowest confidence in their most probable labeling.
- *Margin (M)* [24]: select the $K$ nodes amongst all the nodes in the unlabeled pool that have the lowest margin between their two most probable labellings.
- *Expected Change (EC)* [37]: select the $K$ nodes that has the highest expected number of changed labels.
- *Random Sampling (RND)*: randomly select a batch of nodes (persons and/or scenes) in all graphs (frames).

The Least Confidence [29] and Margin [24] algorithms are originally proposed for simple prediction but we extended them for our structured prediction task. We examine performance of these methods over iterations of active learning. To enable fair comparison, the same amount of data/annotations are given to all the methods.

We trained a base model with the same initial training set (which is a small portion of the whole training set) for each method, and then grow the labeled training set by actively selecting $K$ annotations per iteration.

## 5. Results and Analysis

Below, we compare our method quantitatively to baselines and perform analysis of the results.

### 5.1. Quantitative Results

**Volleyball Dataset**: Figures 4a and 4b show the results of our method and baselines on the Volleyball Dataset. Our proposed method exhibits similar performance, focusing on scene labeling performance first, outperforming all baseline methods at this task and eventually outperforming all at action labeling as well.

**Collective Activity Dataset**: Figures 4c and 4d illustrate the comparison of our method against baselines on the Collective Activity Dataset. Note that performance on both individual action and group activity (scene) labeling are measured. Our method tends to focus its initial effort on labeling scenes, likely because of the gains in structured prediction performance that can be obtained in this manner. Overall, our method outperforms all baselines on this dataset on scene labeling and after a few iterations outperforms all baselines on action labeling as well.

The full details of Figure 4 and a table of the accuracy that each method has achieved at each iteration is reported in the supplementary material.

### 5.2. Comparison to Supervised Methods

For these datasets alternate approaches use all available data supervised. Our experiments show that our method can save ≈ 70% of annotation cost yet achieve similar accuracy as the state-of-the-art supervised methods.

In terms of fully supervised methods, Deng et al. [5] achieved 74.0% *scene accuracy* on the whole training

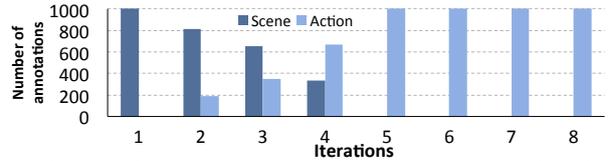
(a) Number of scene/action annotations selected at each iteration for volleyball dataset experiment, in which 1000 annotations are added at each iteration.

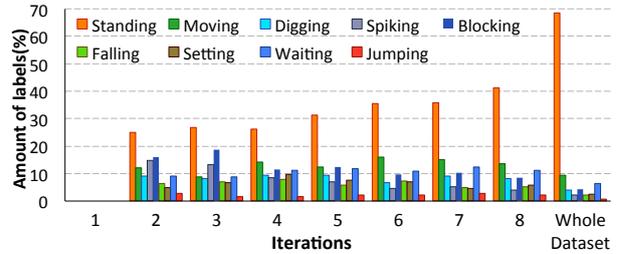
(b) Amount of individual **action** labels chosen at each iteration for volleyball dataset. No action labels are added at iteration 1.

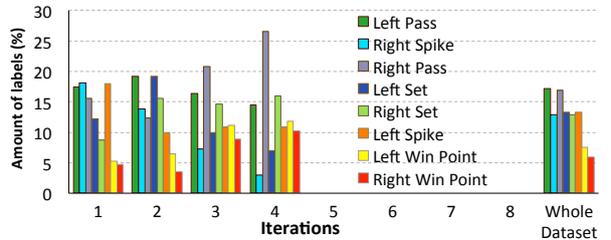
(c) Amount of **scene** labels chosen at each iteration for volleyball dataset. After iteration 4 no scene labels are added.

| | Epochs | Number of **Added** Annotations | | | | | | | | |
|---|---|---|---|---|---|---|---|---|---|---|
| | | 0 | 1000 | 2000 | 3000 | 4000 | 5000 | 6000 | 7000 | 8000 |
| Scene | 90 | 42.4 | 56.9 | 60.6 | 62.5 | 64.4 | 65.3 | 66.1 | 66.1 | 66.2 |
| | 60 | 42.4 | 55.7 | 59.9 | 62.4 | 64.3 | 65.2 | 65.4 | 65.3 | 65.8 |
| | 30 | 42.4 | 53.7 | 58.5 | 61.7 | 62.9 | 64.1 | 64.6 | 64.9 | 65.1 |
| | 15 | 42.4 | 53.2 | 58.4 | 60.8 | 62.6 | 63.3 | 63.4 | 63.6 | 63.8 |
| Action | 90 | 68.1 | 68.4 | 70.7 | 74.3 | 76.1 | 76.8 | 77.3 | 77.7 | 78.1 |
| | 60 | 68.1 | 68.3 | 70.5 | 73.3 | 75.8 | 76.5 | 76.8 | 77.3 | 77.5 |
| | 30 | 68.1 | 68.3 | 70.1 | 72.9 | 74.8 | 75.6 | 76.3 | 76.7 | 76.9 |
| | 15 | 68.1 | 68.2 | 69.7 | 72.5 | 74.1 | 74.3 | 75.4 | 75.7 | 75.8 |

Table 1: Results of our method with different number of epochs at each iteration.

set (11,734 annotations) using the "untied version" of their method. Our method achieves comparable accuracy (74.8%) after 3 iterations of adding 500 annotations (in total 1500 extra annotations) (Table 1 right-side in the supplementary material). In other words, using 33% of the data the same accuracy is achieved. Higher accuracy (76%) is obtained after labeling ≈ 44% of the data. This shows the potential for active learning-based methods for this task.

Ibrahim *et al*. [10] obtain 68.1% accuracy (the non-temporal model) using 44,294 training annotations. Using much less training data, e.g. ≈35% of the data we can achieve results comparable to this fully supervised method.

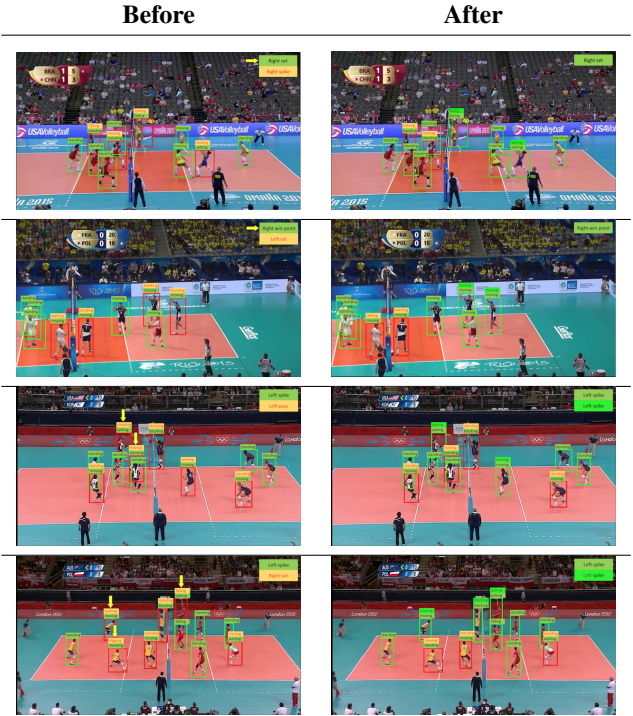

Figure 6: Examples of chosen variables in the Volleyball dataset. Each row shows one pair of before/after an active learning iteration. Left image shows scene with current labels (ground truth in green box, incorrectly predicted labels in yellow box). Yellow arrow shows variable chosen for labeling by oracle by our method. Note that selection of group activity (*scene* annotation) in top 2 rows or actions of specific people (bottom 2 rows) help in correcting other labels afterwards.

### 5.3. Analysis

**Analysis of Nodes Chosen**: Figure 5a shows the number of *scene* and *action* nodes that have been selected by our active learner at each iteration. In the first few iterations, mostly scene nodes have been chosen for labeling and later on action nodes are selected. This makes intuitive sense given the impact that correct scene labels can have on the action labels in a video frame. The reason that our method doesn't outperform baselines in terms of action accuracy at first is that almost no new action annotations are provided to the learner. But after choosing action nodes our method outperforms the baselines. The distributions of labeled action and scene classes at each iteration are illustrated in Figures 5b and 5c respectively. For the sake of comparison, the distribution of different classes in the whole training set is shown in the right side of Figures 5b and 5c.

There is an interesting observation in Figure 5b that 70% of the action labels belong to the *standing* action, which is ≈28,000 annotations. Looking at the distribution of the selected action labels in the first few iterations, the ratio of the *standing* action to other actions is significantly lower than its frequency in the dataset. This indicates that our algorithm considered this class as an easy class and a smaller number of instances were selected. On the other hand, only 1% of the training instances belong to the *jumping* class. Therefore the probability of choosing a sample from this class is below 1%. Nevertheless, our method has selected significantly more from the *jumping* class at each iteration, in order to model the relative complexity of this class.

Regarding the *standing* class, because the number of standing people is enormous compared to other classes, the probability of choosing this class is still higher; thus at each iteration more *standing* instances have been chosen compared to other classes such as *jumping*.

**Effect of number of epochs in training**: In this section we study how the number of epochs used in training affects the accuracy. For a given annotation set often the more epochs of stochastic gradient descent the better accuracy gets. There exist different work-flow patterns for acquiring labels from an oracle (e.g. Amazon Mechanical Turk). These include recruiting workers at sparse intervals, interleaving human labeling with model training, and concurrent training and labeling.

In a deep learning framework, usually with the right choice of learning rate and weight decay, the solver achieves higher accuracies if *enough time* is spent on training. However, in active learning, there might be some cases where user wait time is constrained and the interaction time between a user and the active learner is limited. In this case, less time should be spent on training[2]. We have run experiments that show the performance of our method for different numbers of epochs of learning in between each active data labeling cycle. We have reported our results in Table 1.

In terms of running time, the bottleneck of our pipeline is the training part not the instance selection part. In each iteration of our experiments on the Volleyball Dataset, training for 60 epochs takes 1 hour (only ≈ 5 minutes for sample selection and ≈ 55 minutes for training) on a GeForce GTX 1080 in a Caffe [13] implementation. There is an application-dependent trade-off based on worker recruitment, ramp-up, labeling time, etc. The results that we have provided for different numbers of epochs (running time) in Table 1, illustrate this accuracy-time trade-off.

### 6. Conclusion

In this paper, we presented an active learning approach for partially labeled structured prediction models that can be represented using graphical models. In our method, the initial models are built using a small training subset. Then, in an iterative manner we select a subset of people/scene labels to relearn better models. The selection strategy is based on a novel expected information gain criterion, which is computed from expected entropy reduction. Results demonstrate that our algorithm can improve upon baseline active learning approaches and achieve results competitive with fully supervised methods while using only a fraction of the labeled data.

---
[2]This depends on the setting, one could also use crowdsourced services where the "wait time" might involve different labelers and not be relevant.

# 7. Supplementary Material

## 7.1. Training details

**Learning Rate and Weight Decay Policy**: Our strategy involves continually fine-tuning a network given iterations of new labeled data. Choosing the right learning rate is an important consideration. At every iteration of our active learning algorithm, the solver starts with a base learning rate and decreases it gradually. Initializing the solver with the same learning rate at the beginning would result in low performance since the model has already been fine-tuned using some data. So in our policy, at each iteration, the solver is automatically initialized with a base learning rate that is smaller than the previous iteration's base learning rate by a fixed factor. In order to avoid over-fitting to the smaller amount of data in the beginning, we designed a weight decay policy that the solver at first is initialized with a larger weight decay and at each iteration it decreases by a constant factor. In our experiments we used $0.5$ for the base learning rate multiplication factor and $0.1$ for weight decay.

## 7.2. Explanation of the first stage

The first stage (learner/classifier) of our active learning method could be any structured prediction learning algorithm that involves inference on a graphical model. For the implementation, we used the work of Deng *et al.* [5] as the learner/classifier. The reason is two fold: (1) it is state of the art; (2) the code is publicly available and is a good fit for our problem. For better understanding of our code/method, here we briefly explain [5].

Deng *et al.* [5] develop a method for structured prediction within deep networks. The approach implements the message passing algorithm for graphical model inference in a deep neural network framework. Their framework has two parts (see Fig. 7). The first consists of individual predictions for each output of the model. In the case of activity recognition, these would correspond to predicting the action label for each person as well as the overarching activity label for the scene. Convolutional neural networks (AlexNet [17]) are used for these individual predictions. With two different types of output node (action / scene), two different neural networks are trained, A-CNN an S-CNN. These two networks will be used later for extracting features from the frame and bounding boxes of the people in the scene.

The second part is an inference machine that uses the outputs from these individual neural networks to refine estimates of labels. This second part is akin to a message passing algorithm to conduct inference in a graphical model. Parameters in these messages control how much support / conflict there is between different labels in a scene, similar to potential functions in a graphical model. These parameters are represented as weights in a neural network and can be learned by back-propagation.

**Network parameters**. The parameters that we used for A-CNN and S-CNN of the first stage of our algorithm are as follow. We set the initial base learning rate of A-CNN and S-CNN to $0.0005$ and $0.00005$ respectively; and we set the initial weight decay of both to $0.05$. In the experiment section we explained that we start with large base learning rate and weight decay and gradually reduce it at each iteration. Based on our heuristic approach, we multiply the base learning rate by $0.2$ at each iteration. Similar to learning rate, we multiply the weight decay by $0.2$ at each iteration. Then the model is initialized with the new parameters and learning continues. The batch sizes that we used for A-CNN and S-CNN are $600$ and $384$ respectively. As for the second stage we set the initial base learning rate to $0.002$ and at every five iteration we multiply it by $0.5$. However, for second stage we used the same weight decay and multiplier that we used for A-CNN and S-CNN.

### 7.2.1 Experiments

We have conducted two sets of experiments on two different datasets. The plots in Fig. 4 of the paper illustrates all the comparison of our method to the baselines on Collective Activity Dataset (Table 2a) and Volleyball Dataset (Table 2b). These plots are based on the numbers in the Table 2. The tables on the left and on the right show the results of adding $K = 1000$ and $K = 500$ annotations per iteration, respectively. The first column of the tables shows the accuracy of the trained model on the initial training set, which is same for all the methods. Subsequent columns show the results of the iterations of active learning.

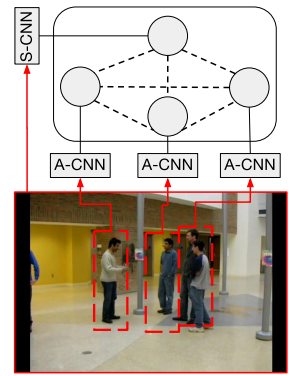

Figure 7: Overview of the Stage 1 learner. Bounding boxes of individuals and the entire frame are passed through two CNNs to predict their labels. This information is then passed to a Recurrent Neural Network (called BP-RNN). The BP-RNN simulates the belief propagation algorithm on the graphical model of the scene, in which there are nodes corresponding to the overarching scene label and action label of every person.

### (a) Collective Activity Dataset

**Left: 1000 per iteration**

| | Method | 0 (0%) | 1000 (≈9%) | 2000 (≈17%) | 3000 (≈26%) | 4000 (≈35%) | 5000 (≈44%) | 6000 (≈53%) |
|---|---|---|---|---|---|---|---|---|
| Scene | SA [20] | 69.8 | 71.8 | 71.6 | 71.8 | 72.1 | 71.6 | 71.6 |
| | M [24] | 69.8 | **72.3** | **72.3** | 72.3 | 69.7 | 71.3 | 72.1 |
| | LC [29] | 69.8 | 70.8 | 70.1 | 71.0 | 71.4 | 71.2 | 71.7 |
| | EC [37] | 69.8 | 71.6 | 72.3 | 71.8 | 71.7 | 72.2 | 72.2 |
| | RND | 69.8 | 67.6 ±2.7 | 68.5 ±2.2 | 70.8 ±1.5 | 71.4 ±1.4 | 72.3 ±1.5 | 72.7 ±1.3 |
| | Ours | 69.8 | 71.0 | 70.4 | **73.5** | **75.7** | **76.2** | **76.2** |
| Action | SA [20] | 66.5 | 69.1 | **72.0** | **72.3** | 72.1 | 72.3 | 72.3 |
| | M [24] | 66.5 | **70.2** | 70.1 | 70.5 | 69.4 | 70.0 | 70.1 |
| | LC [29] | 66.5 | 68.1 | 68.5 | 69.6 | 69.4 | 69.8 | 70.3 |
| | EC [37] | 66.5 | 69.1 | 69.4 | 69.9 | 69.5 | 70.3 | 70.6 |
| | RND | 66.5 | 67.8 ±1.4 | 67.5 ±1.0 | 68.2 ±0.8 | 71.6 ±1.1 | 71.4 ±1.0 | 72.1 ±0.6 |
| | Ours | 66.5 | 69.5 | 71.2 | 71.1 | **72.5** | **73.3** | **73.2** |

**Right: 500 per iteration**

| | Method | 0 (0%) | 500 (≈4%) | 1000 (≈9%) | 1500 (≈13%) | 2000 (≈17%) | 2500 (≈22%) | 3000 (≈26%) |
|---|---|---|---|---|---|---|---|---|
| Scene | SA [20] | 69.8 | 70.2 | 71.2 | 70.2 | 70.3 | 70.3 | 70.3 |
| | M [24] | 69.8 | 71.4 | 72.4 | 72.4 | 70.8 | 72.3 | 71.9 |
| | LC [29] | 69.8 | 70.3 | 70.6 | 71.1 | 70.9 | 71.3 | 71.4 |
| | EC [37] | 69.8 | 70.9 | 71.5 | 72.2 | 72.0 | 71.9 | 72.0 |
| | RND | 69.8 | 68.7 ±1.3 | 68.9 ±1.2 | 69.0 ±0.9 | 69.2 ±0.9 | 69.6 ±1.0 | 70.1 ±0.8 |
| | Ours | 69.8 | **72.6** | **73.2** | **75.4** | **74.8** | **76.0** | **75.2** |
| Action | SA [20] | 66.5 | 68.2 | **70.9** | **71.5** | 71.3 | 71.6 | 71.6 |
| | M [24] | 66.5 | **69.2** | 69.9 | 70.5 | 70.2 | 70.3 | 70.0 |
| | LC [29] | 66.5 | 68.2 | 68.4 | 68.4 | 69.5 | 70.2 | 69.9 |
| | EC [37] | 66.5 | 68.5 | 69.3 | 69.8 | 69.9 | 70.1 | 70.0 |
| | RND | 66.5 | 69.4 ±1.1 | 69.7 ±0.9 | 69.7 ±0.8 | 70.1 ±0.6 | 70.2 ±0.7 | 69.8 ±0.7 |
| | Ours | 66.5 | 68.7 | 70.7 | 71.2 | **72.2** | **72.1** | **72.6** |

(a)

### (b) Volleyball Dataset

**Left: 1000 per iteration**

| | Method | 0 (≈0%) | 1000 (≈2%) | 2000 (≈4%) | 3000 (≈6%) | 4000 (≈9%) | 5000 (≈11%) | 6000 (≈13%) | 7000 (≈16%) | 8000 (≈18%) |
|---|---|---|---|---|---|---|---|---|---|---|
| Scene | SA[20] | 42.4 | 53.4 | 56.5 | 57.8 | 58.6 | 58.8 | 59.1 | 59.6 | 59.9 |
| | M[24] | 42.4 | 48.9 | 49.9 | 50.1 | 50.2 | 50.4 | 50.6 | 51.1 | 51.2 |
| | LC[29] | 42.4 | 45.4 | 46.2 | 46.3 | 46.8 | 47.0 | 46.7 | 46.8 | 46.8 |
| | EC [37] | 42.4 | 46.1 | 47.1 | 47.8 | 49.2 | 49.1 | 50.7 | 50.8 | 51.1 |
| | RND | 42.4 | 44.7 ±2.3 | 46.6 ±1.2 | 47.5 ±1.1 | 48.8 ±1.4 | 49.7 ±1.1 | 49.8 ±1.0 | 50.0 ±0.8 | 50.1 ±0.8 |
| | **Ours** | 42.4 | **55.7** | **59.9** | **62.4** | **64.3** | **65.2** | **65.4** | **65.3** | **65.8** |
| Action | SA[20] | 68.1 | 69.1 | 70.4 | 71.9 | 72.4 | 74.7 | 75.1 | 75.4 | 75.5 |
| | M[24] | 68.1 | **70.9** | **71.3** | 71.6 | 72.1 | 74.1 | 75.2 | 75.9 | 76.1 |
| | LC[29] | 68.1 | 70.1 | 70.5 | 71.0 | 71.2 | 71.5 | 72.0 | 72.0 | 72.1 |
| | EC [37] | 68.1 | 68.9 | 69.6 | 72 | 72.2 | 72.9 | 73.7 | 73.8 | 74.7 |
| | RND | 68.1 | 68.8 ±1.1 | 70.1 ±1.3 | 70.9 ±1.2 | 71.6 ±1.0 | 72.1 ±0.9 | 72.3 ±0.9 | 72.4 ±0.8 | 72.5 ±0.8 |
| | **Ours** | 68.1 | 68.3 | 70.5 | **73.3** | **75.8** | **76.5** | **76.8** | **77.3** | **77.5** |

**Right: 500 per iteration**

| | Method | 0 (≈0%) | 500 (≈1%) | 1000 (≈2%) | 1500 (≈3%) | 2000 (≈4%) | 2500 (≈5%) | 3000 (≈6%) | 3500 (≈8%) | 4000 (≈9%) |
|---|---|---|---|---|---|---|---|---|---|---|
| Scene | SA[20] | 42.4 | 51.4 | 53.4 | 54.7 | 55.0 | 55.2 | 56.1 | 56.3 | 56.4 |
| | M[24] | 42.4 | 46.4 | 47.5 | 49.7 | 50.1 | 50.2 | 50.1 | 51.3 | 51.0 |
| | LC[29] | 42.4 | 44.9 | 46.4 | 46.7 | 46.5 | 47.0 | 47.1 | 46.8 | 47.0 |
| | EC [37] | 42.4 | 45.6 | 46.2 | 46.9 | 47.1 | 47.2 | 48.1 | 48.0 | 48.9 |
| | RND | 42.4 | 44.1 ±2.5 | 46.0 ±1.4 | 46.8 ±1.2 | 47.5 ±1.3 | 48.3 ±1.2 | 48.7 ±0.9 | 48.9 ±0.8 | 49.1 ±0.7 |
| | **Ours** | 42.4 | **53.7** | **58.4** | **60.7** | **62.0** | **62.0** | **62.3** | **62.9** | **62.9** |
| Action | SA[20] | 68.1 | 69.0 | **70.7** | 71.4 | 72.1 | 72.9 | 74.1 | 74.6 | 74.9 |
| | M[24] | 68.1 | **70.1** | 70.5 | 72.1 | 72.4 | 73.1 | 73.8 | 74.3 | 75.0 |
| | LC[29] | 68.1 | 68.5 | 68.9 | 69.3 | 71.1 | 71.1 | 71.3 | 71.4 | 71.8 |
| | EC [37] | 68.1 | 68.7 | 69.6 | 71.2 | 71.6 | 71.5 | 71.9 | 72.0 | 72.1 |
| | RND | 68.1 | 68.6 ±1.5 | 69.0 ±1.4 | 70.0 ±1.2 | 70.6 ±1.2 | 70.9 ±1.0 | 71.2 ±0.9 | 71.6 ±0.8 | 71.9 ±0.9 |
| | **Ours** | 68.1 | 68.3 | 69.0 | **72.2** | **72.8** | **73.9** | **74.6** | **75.4** | **76.3** |

(b)

Table 2: Results of comparison of our method against baselines on a) *Collective Activity Dataset* and b) *The Volleyball Dataset*. The numbers in the table are accuracies of the **action** and **scene** labels(%). For all the methods we start from same small initial labeled set so the accuracies of the first column are exactly the same. For each dataset, two sets of experiments are conducted that differ in number of annotations added at each iteration. Tables on the left and on the right report results of experiments with 1000 and 500 number of annotations added per iteration, respectively. For all the experiments, models are trained for 60 number of epochs at each iteration